\documentclass{elektr}
\usepackage{hyperref}
\hypersetup{
colorlinks=true,
urlcolor=blue,
citecolor=blue}
\usepackage[all]{xy,xypic}
\usepackage{amsfonts,amssymb,amsmath,amsgen,amsopn,amsbsy,theorem,graphicx,epsfig}
\usepackage{eufrak,amscd,bezier,latexsym,mathrsfs,eurosym,enumerate}
\usepackage[utf8]{inputenc}
\usepackage[english]{babel}
\usepackage{cleveref,multirow}
\usepackage[dvipsnames]{xcolor}
\usepackage[pagewise]{lineno}
\yil{}
\vol{}
\fpage{}
\lpage{}
\doi{}

\title{Neural relation extraction: a survey}

\author[AUTHOR and AUTHOR and AUTHOR]{
\textbf{Mehmet AYDAR$^{1,2}$, {O}zge BOZAL$^{1,3}$\thanks{ozge.bozal@huawei.com}~, Furkan OZBAY$^{1,4}$}\\
$^{1}$AI Enablement Department, Huawei Turkey Research and Development Center, Istanbul, Turkey\\
$^{2}$Enterprise Architecture and Technology Innovation, Ford Otosan, Istanbul, Turkey\\
$^{3}$Department of Computational Science and Engineering, Bogazici University, Istanbul, Turkey\\
$^{4}$Department of Computer Engineering, Yildiz Technical University, Istanbul, Turkey
\\ [1.8em]

\rec{.201}
\acc{.201}
\finv{..201}
}

\def\E{\ifmmode{\mathbb E}\else{$\mathbb E$}\fi} %natural numbers
\def\N{\ifmmode{\mathbb N}\else{$\mathbb N$}\fi} %natural numbers
\def\R{\ifmmode{\mathbb R}\else{$\mathbb R$}\fi} %real numbers
\def\Q{\ifmmode{\mathbb Q}\else{$\mathbb Q$}\fi} %rational numbers
\def\C{\ifmmode{\mathbb C}\else{$\mathbb C$}\fi} %complex numbers
\def\H{\ifmmode{\mathbb H}\else{$\mathbb H$}\fi} %complex numbers
\def\Z{\ifmmode{\mathbb Z}\else{$\mathbb Z$}\fi} %integers
\def\P{\ifmmode{\mathbb P}\else{$\mathbb P$}\fi} %real numbers
\def\T{\ifmmode{\mathbb T}\else{$\mathbb T$}\fi} %real numbers
\def\SS{\ifmmode{\mathbb S}\else{$\mathbb S$}\fi} %real numbers
\def\DD{\ifmmode{\mathbb D}\else{$\mathbb D$}\fi} %real numbers

\newcommand{\bse}{\begin{subequations}}
\newcommand{\ese}{\end{subequations}}
\newcommand{\ben}{\begin{enumerate}}
\newcommand{\een}{\end{enumerate}}
\newcommand{\bens}{\begin{enumerate*}}
\newcommand{\eens}{\end{enumerate*}}
\newcommand{\be}{\begin{equation}}
\newcommand{\ee}{\end{equation}}
\newcommand{\bea}{\begin{eqnarray}}
\newcommand{\eea}{\end{eqnarray}}
\newcommand{\baa}{\begin{eqnarray*}}
\newcommand{\eaa}{\end{eqnarray*}}
\newcommand{\bc}{\begin{center}}
\newcommand{\ec}{\end{center}}

\theoremstyle{corollary}

\theoremstyle{lemma}

\theoremstyle{proposition}

\theoremstyle{axiom}

\theoremstyle{conjecture}

\theoremstyle{example}

\theoremstyle{definition}
\theoremstyle{remark}
\begin{document}

\maketitle

\begin{abstract} Neural relation extraction discovers semantic relations between entities from unstructured text using deep learning methods. In this study, we present a comprehensive review of methods on neural network based relation extraction. We discuss advantageous and incompetent sides of existing studies and investigate additional research directions and improvement ideas in this field.

\end{abstract}

\maketitle

\section{Introduction}
\label{intro}
Never-ending information generation and sharing on the Web provides us with abundant data, most of which constitute the unstructured text sources. To better make sense of and draw associations among those data, we, human beings, use relational facts among the subjects (entities) in the text. For a more comprehensive understanding of specific domains such as bioinformatics, finance, social networking etc., we need computers to process those information. 

It is essential to represent the information delivered by the text in machine-readable format. One way to do that is to represent entities and their relations in so called triples, which indicate unambiguous facts about entities. A triple ($h$, $r$, $t$) implies that entity $h$ has relation $r$ with another entity $t$.  Knowledge graphs (KG) such as FreeBase \citep{bollacker2008freebase} and DBpedia \citep{auer2007dbpedia} are examples of such representations. They are directed and labeled graph structured data which aim to express such explicit semantics and relations of entities in triple form.

Relation extraction is a sub-task of natural language processing (NLP) which aims to discover relations $r$ between entity pairs $h$ and $t$ given unstructured text data. Earlier work on relation extraction from text heavily relies on kernel based and feature based methods \citep{pawar2017relation}.However, recent research studies make use of data-driven deep learning methods eliminating conventional NLP approaches for relation extraction. \citet{kumar2017survey} explained how the conventional deep learning methods are integrated into relation extraction. \citet{smirnova2018relation} reviewed relation extraction literature focusing on distant supervision. As the number of research studies on relation extraction increases, the need of a survey on current state-of-the-art of neural relation extraction methods arises.

This work provides a comprehensive and comparative review on the research field, focusing on the challenges together with improvement ideas. 
Section \ref{approach} explains various approaches for relation extraction. In section \ref{type} neural relation extraction methods are classified in terms of data supervision and explained. Section \ref{challenges} describes existing challenges in this field of research.  In section \ref{dataset}, commonly used datasets in model assessment are evaluated. We discuss possible future research directions and improvement ideas in section \ref{ideas} and we conclude our survey in section \ref{conclusion}.

\section{Relation extraction approaches}
\label{approach}
In this section, we categorize neural relation extraction methods regarding their assumptions on expressiveness of training instances about the relations. 

\subsection{Sentence-level relation extraction}
\label{sl-re}
In this approach, sentence-based annotated training data is used. Annotation contains sentence-triple alignment information, such that sentences in the training set are labeled with the triples. 
Once trained, the model's objective is to predict new relations given new entity pairs.
However, insufficient amount of training data is a major drawback as labeled data is not always available in real life scenarios. Table \ref{tab:sentence-level-dataset} shows total number of relations and sentences provided in common relation extraction datasets according to OpenNRE framework \citep{han2019opennre}. 

\begin{table}[h!]
\caption{Number of relations and sentences provided in common relation extraction datasets.}
\begin{center}
\begin{tabular}{ |c|c|c|c| } 
\hline
\textbf{Dataset} & \textbf{\#relations} & \textbf{\#sentences} \\
\hline
SemEval-2010 Task 8 & 9 & 6,647 \\ 
TACRED  & 42 & 21,784 \\ 
Wiki80 & 80 & 56,000 \\ 
\hline
\end{tabular}

\label{tab:sentence-level-dataset}
\end{center}
\end{table}

\subsection{Bag-level relation extraction}
\label{bag-re}
Since labeling data in deep learning requires a lot of manual effort, external knowledge bases are used to enhance weakly labeled training set. Knowledge graphs contain information regarding the relations between the entities in the form of ($head$, $relation$, $tail$) triples. For creating distant supervision datasets such as NYT, entity pairs in a triple are aligned with sentences that contain $head$ and $tail$ entities in the natural text. In this approach, the sentences matched by an entity pair constitute a bag. For this reason these datasets are noisy. Besides that, they are imbalanced, that is, the instances are not evenly distributed across relations.
 
There are different selection methods to weigh the expressiveness of a bag's instances. One might choose a maximum, average or attention selector which considers the most relevant instance, all instances or weighted average of all instances, respectively \cite{riedel2010modeling, hoffmann2011knowledge, lin2016neural}. More details regarding this approach is given in section \ref{distant-supervision-type}.

\subsection{Document-level relation extraction}
\label{document-re}	
Sentence-level approach lacks in grasping entity pair relations across a document \citep{yao2019docred, quirk2017distant}, that is to say, it ignores relations which can be deduced only by understanding several sentences within a document. This can especially be vital for some domains such as drug-side effect relations in pharmaceutical documents \citep{verga2018simultaneously}. The study of \citet{quirk2017distant} is the first to address this problem in distantly supervised setups and proposes a document-level graph representation to extract more relations. DocRED \citep{yao2019docred} provides a benchmark dataset for document-level relation extraction which contains relations that can only be extracted from multiple sentences. As of today, performance of document-level relation extraction methods fall behind the human performance when it comes to cross-sentence reasoning, therefore this approach needs more effort.

\section{Types of relation extraction}
\label{type}
\subsection{Supervised relation extraction}
\label{supervised}
Supervised neural relation extraction from text uses sentence-level relation extraction approach which requires labeled relation-specific training data. Many of the studies on this task rely on classifying entity pairs according to the particular relations they are assigned to. We list the results of existing methods in Table \ref{table:1}.

\subsubsection{Conventional neural models for relation extraction}
Recent research studies focus on extracting relational features with neural networks instead of manual work \citep{socher2012semantic, zeng2014relation, santos2015classifying}. \citet{socher2012semantic} proposes a recurrent deep neural network model which admits a compositional vector representation of words and phrases on a parse tree. Each expression is represented by both a vector and a matrix, the former encodes semantic information of an expression and the latter encodes how much it influences the meaning of syntactically neighboring expressions.

In relation classification, drawing global features of relations within a sentence is a crucial task. Accordingly, \citet{zeng2014relation} utilize convolutional neural networks which can combine local features to obtain a globally representative one.
To decrease effects of undesirable artificial classes of relations in prediction, \citet{santos2015classifying} introduces a convolutional deep learning model that admits a pairwise ranking loss function and achieves better results than the former model.

TACRED is introduced by \citet{zhang2017position}, a relation extraction dataset created based on yearly TAC KBP evaluations. Proposed LSTM sequence model coupled with entity position-aware attention mechanism trained on TACRED outperforms the TAC KBP 2015 slot filling system.

In \citet{zhang2015relation}'s work, it is claimed that RNN based relation extraction models excel CNN based models, for the reason that CNNs can only capture the local features, whereas RNNs are capable of learning long-distance dependency between entities.

An LSTM model proposed by \citet{xu2015classifying} takes advantage of the shortest dependency path (SDP) between entities. They claim that the words along SDP are more informative. Dependency trees are directed graphs, therefore, there is the need of differentiating whether the first entity is related to the second entity or the relation implies the reverse direction. For this purpose, the SPD is divided into two sub-path, each is directed from the entity towards the ancestor node.

Uni-directional LSTM models lack in expressing the complete sequential information of the sentences. \citet{zhang2015bidirectional} use bidirectional LSTM model (BLSTM) to better represent the sentences.

Meaningful information can be located anywhere in the sentence. Instead of using features from lexical sources such as dependency parsers and named entity recognizers, \citet{zhou2016attention} incorporate attention mechanism to BLSTM network to capture more informative parts of the sentence.

Pipeline approaches which first find the entities than match them with the appropriate relations are prone to error-propagation, namely, errors in the first part can't be alleviated in the relation classification part. Recent models study extraction of entities together with their relations. \citet{wei2019novel} introduces a hierarchical tagging scheme that maximizes the likelihood of input data and the relational triples. Given a sentence, first it finds the subjects, then for each relation $r$, it tags the appropriate objects, which can also be an empty set. This way, multiple triples can be extracted.

\subsubsection{Pre-trained language models for relation extraction}
\label{pre-trained}
Transfer learning is commonly used in deep learning to transfer existing knowledge of a model of a specific task to another similar or related task's model. The former models are called pre-trained models and they save plenty of time and computational power. For NLP tasks there are several widely used pre-trained models such as BERT \citep{devlin2018bert}, Transformer-XL \citep{dai2019transformer} and OpenAI's GPT-2 \citep{radford2019language}. 

Commonly preferred pre-trained language model in relation extraction studies, BERT, is an unsupervised transformer which is trained to predict the next sentence given a sequence of sentences, and also for masked language model. BERT's model captures the contextual information of a word in a given sentence, along with the semantic relation of a sentence to the neighboring sentences in building the whole text. \citet{wu2019enriching} adjust the pre-trained BERT model to handle both sentence and its entities and they achieved better results on SemEval-2010 task 8 dataset than other conventional deep learning methods. \citet{soares2019matching} aim to build task agnostic, efficient relation representations from natural text using BERT. They achieved better results than previous models on SemEval-2010 task 8 and other models trained on TACRED. \citet{zhao2019improving} achieved the best result on SemEval-2010 task 8. They extract graph topological features on top of BERT embeddings. On the other hand, \citet{wei2019novel}, which also utilizes BERT, achieved best results on the distantly supervised NYT dataset.

\begin{table}[h!]
\caption{States F1 scores of supervised relation classification approaches using Semeval 2010 Task-8 as input dataset.}
\begin{center}
%\resizebox{\columnwidth}{!}{
\begin{tabular}{c|c|c} 
 \hline
 & & \textbf{SemEval 2010} \\ [0.5ex]
 \hline\hline  \multirow{4}{*}{CNN-based}
%   \rowcolor{lightgray}
 & \citet{zeng2014relation}& 82.7  \\ 
& \citet{nguyen2015relation}&  82.8  \\
% \rowcolor{lightgray}
& \citet{santos2015classifying} & 84.1 \\
& \citet{wang2016relation} & 88.0 \\
 \hline\hline
 \multirow{2}{*}{RNN-based}
 & \citet{zhou2016attention} &84.0 \\
 & \citet{cai2016bidirectional} &86.3 \\
 \hline\hline
 \multirow{4}{*}{BERT-based}
 & \citet{wei2019novel} &87.5 \\
 & \citet{soares2019matching} &89.2  \\
 & \citet{wu2019enriching} &89.25 \\
 & \citet{zhao2019improving} &90.2 \\
 \hline
\end{tabular}
%}
\label{table:1}
\end{center}
\end{table}

\subsection{Relation extraction using distant supervision}
\label{distant-supervision-type}
Distant supervision aligns triples in a related knowledge graph with the sentences in input text, in order to automatically generate training data. 
Distant supervision assumes the responsibility to determine which sentence supports which relation and to what degree it expresses the relation of interest. In other words, distant supervision labels sentences with appropriate relations, and generates an error-prone training set consisting of possibly wrong labeled instances, which in turn is used to train relation extraction models. 

\citet{mintz2009distant} is the first to use this technique. The assumption was that given a triple from the knowledge graph, all sentences that contain the head and tail entities of the triple express the corresponding relation. As a matter of fact, it causes wrong-labeling problem. For instance, consider a triple {\fontfamily{qcr}\selectfont (Bill\_gates, Founder, Microsoft)} from knowledge base and two sentences below: 
\\
{\fontfamily{qcr}\selectfont ``Bill Gates is the co-founder of Microsoft.''} and \\
{\fontfamily{qcr}\selectfont ``The greatest mistake of Bill Gates cost Microsoft \\ 
\$400 billion.''}\\
It is clear that the first sentence expresses {\fontfamily{qcr}\selectfont Founder} relation, as the latter does not. Therefore, the training set including the second sentence is said to be noisy or wrongly labeled. Subsequent studies in distant supervision utilized the same trivial idea of triples in sentence alignment as the original paper. However, they differ in the machine learning models and feature encoders along with their approach to sentence labeling with appropriate relations and solving the wrong labeling problem caused by distant supervision. Reader can find the results of existing distantly supervised methods in Table \ref{table:2}.

\subsubsection{Sentence-triple alignment}

Four different frameworks exist for labeling sentences with appropriate relations, which are \textit{single-instance single-label (SISL)}, \textit{multi-instance single-label (MISL)}, \textit{single-instance multi-label (SIML)} and \textit{multi-instance multi-label (MIML)} learning. Regarding distant supervision, an instance refers to a sentence in the natural text and a label refers to a relation captured by the knowledge base. 
Single-instance models assume that a particular relation is derived from only one sentence, while multi-instance approach admits more than one sentence to represent a relation. Multi-instance learning is the bag-level distant supervision approach as explained in section \ref{bag-re}.
In single-label methods a particular sentence is relevant to only one relation, whereas in multi-label approach a sentence can express more than one relation. In this sense, MIML learning framework is more realistic, however, it is necessary to employ efficient ranking and denoising strategies.  

Early methods used conventional NLP methods like dependency parsing and POS tagging for denoising strategies in distant supervision. \citet{riedel2010modeling} assume multi-instance approach which is prone to producing noisy labels for training, in case none of the input sentences expresses the relation. \citet{hoffmann2011knowledge} and  \citet{surdeanu2012multi} proposed (multi-instance, multi-label) learning to cover the overlapping triples problem. However, conventional NLP based methods suffer from propagation of errors generated by NLP tools.

Later studies have relied on deep learning methods to solve the wrong-labeling problem in distant supervision.

\subsubsection{Solving wrong-labeling problem with deep learning methods}
\label{noise}
Distant supervision takes on the annotation burden, however it is obliged to wrong labeling problem. Multi-instance learning \citep{dietterich1997solving} aims to relieve problems caused by ambiguously-labeled training data. To denoise training instances of distant supervision, multi-instance learning has become the remedy in relation extraction studies \cite{riedel2010modeling, hoffmann2011knowledge, surdeanu2012multi, zeng2015distant}. \citet{riedel2010modeling} try to remedy wrong labeling with an undirected graphical model. \citet{hoffmann2011knowledge} focus on multi-instance learning with a probabilistic graphical model. Entity pairs in a corpus do not necessarily imply only one relation. In this direction, \citet{surdeanu2012multi} introduce a graphical model with latent variables, which can jointly model the entities and relations in multi-instance multi-label learning fashion.

First neural network model for multi-instance learning with distant supervision was proposed by \citet{zeng2015distant}. 
The method draws relational features with piece-wise convolutional neural network.
The assumption is that, given a relation type, at least one of the input sentences that contain a specific entity pair, is informative, and it considers only the most expressive sentence in training and prediction. Obviously, this method neglects a large amount of data which might also be informative on that relation.

In \citet{lin2016neural}'s work, each sentence is ranked using the attention mechanism based on how well it represents a specific relation. Therefore, it suppresses the noisy ones rooted from distant supervision.
To better extract the most appropriate relations, especially in ambiguous cases, \citet{ji2017distant} formulate entity descriptors to include background information which operate on the instances weighted by sentence-level attention.  

Relations are not individual tags, on the contrary, they are in semantic correlation with each other. To incorporate the rich information covered by relational correlations, \citet{han2018hierarchical} apply a hierarchical attention on each bag of instances.

Another approach is accounting for the information covered by knowledge graphs. \citet{han2018neural} introduce a joint representation learning model for knowledge graph and text, the mutual guidance of which is fed back to the model under an attention mechanism to highlight the significant features of both.
To benefit more from the knowledge graphs, \citet{wang2018label} propose a novel distant supervision approach which refuse the hard labels imposed by regular distant supervision methods, rather, they train the relation classifier directly from KGs with soft labels.
 
Recent research papers \citep{liu2016effective, ratner2016data,  zheng2019diag} confirm that including high quality human annotation brings significant improvement in relation extraction by alleviating the noise. \citet{zheng2019diag} suggest a reinforcement learning based pattern extraction method to ease pattern-writing work for human experts. The pattern-instance pairs are subject to human annotation to be used in fusing the different labeling methods such as distant supervision and relational patterns.

Based on the complementarity and consistency properties of different languages, \citet{lin2017neural} combined mono-lingual and cross-lingual attention to take advantage of both language-specific features and the patterns that bear resemblance across languages. They aggregate sentence encodings with weighted attentions to further use in relation prediction. Sequel to this work, \citet{wang2018adversarial} investigate the effect of incorporating adversarial training in relation extraction. To alleviate possible incompetency in finding consistent patterns across different languages, this work defines a discriminator which can determine the language of each instance.

\begin{table}[h!]
\caption{States mean precision scores of distantly supervised relation classification methods on NYT dataset. ``Held-out'' and`` Manual'' indicates that the scores are from held-out and manual evaluation, respectively.}
\begin{center}
\begin{tabular}{c|c|c} 
 \hline
& & \textbf{NYT}\\ [0.5ex] 
 \hline\hline
 \multirow{4}{*}{Held-out}
& \citet{jiang2016relation} & 72.0  \\
& \citet{lin2016neural} &72.2 \\
& \citet{han2018neural} &71.0 \\
& \citet{han2018hierarchical} &81.6 \\
\hline\hline
\multirow{3}{*}{Manual}
& \citet{zeng2015distant} &78.3 \\
& \citet{ji2017distant} &81.3 \\
&  \citet{wang2018label} &86.9 \\

 \hline
\end{tabular}
\label{table:2}
\end{center}
\end{table}

\subsubsection {Model extensions}
\label{model-textension}
Research studies on relation extraction with distant supervision are not limited to pure deep learning methods. Recent approaches extended their models by incorporating various NLP tools and machine learning methods.

Adversarial learning is used for training classifiers that are robust to both unmodified and perturbed samples. Essentially, it is widely used in supervised learning. \citet{wu2017adversarial} experiment the effects of using adversarial training in relation extraction with distant supervision on convolutional and recurrent neural network architectures and show that it can improve the performance of both.
Redistribution of distantly supervised data can boost the performance of relation classifiers. 
\citet{qin2018dsgan} propose a denoising method for distant supervision relation extraction datasets, in cases where the true positives are more prevalent than the false positives. Similar to generative adversarial networks, this method re-labels the positively labeled instances provided by distant supervision. 
Another approach is accounting for reinforcement learning to handle the noisy instances. \citet{feng2018reinforcement} decompose the relation extraction problem into two tasks: instance selection and relation classification. Instance selector is a reinforcement learning agent which selects the most appropriate instance using the relation classifier's weak supervision. 
To redistribute the distantly supervised data, \citet{qin2018robust} incorporate reinforcement learning, the policy of which is based on the mere classification performance.

Other than the aforementioned methods utilizing adversarial networks and reinforcement learning, there are also other advanced training methods to overcome the drawbacks of distant supervision. For instance, \citet{takamatsu2012reducing} utilize a generative model to predict the wrongly labeled patterns in distant supervision. 

Unlike studies regarding sentence-level denoising, \citet{liu2017soft} adopt entity pair-level denoising approach and derive soft labels for each entity pair bag, which are subject to change during training.
Different from the above studies, the noise-tolerant model introduced by \citet{huang2017deep} utilize deep residual learning \cite{he2016deep}.
\citet{zeng2017incorporating} incorporate path relations from text corpus, namely, they build a model which can handle relations that can be driven from several sentences.

\subsection{Relation extraction using few-shot approach}
\label{fewshot-re}
Few-shot learning is a learning method, in which in contrast to regular deep learning methods the amount of available training data is small. The assumption is that reliable algorithms can be built to achieve competitive performance to the models trained with abundant data. We list some studies related to few-shot learning for relation extraction in Table \ref{table:3}. For the purpose of experimenting few-shot learning algorithms for relation extraction, \citet{han2018fewrel} provide the ``FewRel'' dataset. Prototypical networks \cite{snell2017prototypical}, which admit prototypes rather than classes, are used in few-shot learning scenarios for relation extraction\cite{gao2019hybrid}.
The model proposed by \citet{soares2019matching} outperforms human accuracy on few-shot relation matching.
\citet{ye2019multi} introduce an aggregation network model and a matching mechanism which is multi-level.
    
\begin{table}[h!]
\caption{Accuracy scores of few-shot relation classification methods with the best performing configuration on FewRel dataset.}
\begin{center}
\begin{tabular}{c|c|c} 
 \hline
& \textbf{Best configuration} & \textbf{FewRel} \\ [0.5ex] 
 \hline\hline
  \citet{soares2019matching} & 5 Way 1 Shot &88.9 \\
 \citet{gao2019hybrid} & 5 Way 10 Shot &92.06\\
 \citet{ye2019multi} & 5 Way 5 Shot &92.66 \\
 
 \hline
\end{tabular}
\label{table:3}
\end{center}
\end{table}

\section{Challenges of relation extraction}
\label{challenges}
Challenges in neural relation extraction regarding available data and existing contextual and structural approaches are presented in this section.

\subsection{Overlapping triples}

An entity ({\it SingleEntityOverlap}) or even an entity pair ({\it EntityPairOverlap}) may imply more than one relation in a sentence. Most studies identify entities before the relation classification which assumes that each entity pair is assigned to a single relation (see section \ref{supervised}). 
\citet{zeng2018extracting} propose an end-to-end model which considers relation extraction to be a triple generation problem and applies a copy mechanism to cope with overlapping triples. Another approach proposed by \citet{takanobu2019hierarchical} admits a hierarchy of high-level relation indicator detection to mine the relations in a sentence and low-level entity mention extraction to match these relation to the corresponding entities. GraphRel, introduced by \citet{fu2019graphrel} is a graph convolutional network based neural model that jointly learns entities and relations. It excels the former methods in solving the overlapping triples problem by incorporating the regional and sequential dependency features of words. Unlike the aforementioned methods, \citet{wei2019novel} offer a new formulation for learning relational triples that first identifies subjects, then the relations with a BERT-based subject tagger module, and finally identifies the objects with a relation-specific object module. 

\subsection{Noise in distant supervision}
% \noindent {\bf Noise in distant supervision.} 
Relation extraction needs large amount of annotated data. 
To handle this problem, recent studies incorporate distant supervision which brings its own drawbacks.
Distant supervision faces the problem of wrong labeled sentences troubling the training due to the excessive amount of noise. Related studies try to remedy this problem by sentence-level attention \cite{lin2016neural}, hierarchical attention \cite{han2018hierarchical}, multi-lingual knowledge extraction \cite{lin2017neural}, joint extraction with knowledge graphs \cite{han2018neural} or introducing human annotation to relation extraction \cite{liu2016effective, ratner2016data, zheng2019diag}. %
Detailed information on these methods is given in section \ref{noise}.

\subsection{Few-shot instances}
% \noindent {\bf Few-shot instances.} 
Few-shot based modelling is especially challenging for NLP tasks, since text data is noisy and human annotators tend to be mistaken in language-specific tasks \citep{gao2019hybrid}. \citet{han2018fewrel} investigate few-shot learning for relation extraction and provide a dataset for this specific task. \citet{gao2019fewrel} improve the former dataset by addressing domain adaptation issues and ``none-of-the-above'' case which adds extra class to the model. Prototypical networks which assume classification models built on prototypes rather than class labels enable the classifier to identify new classes when only few instances are present for each of those \citep{snell2017prototypical, gao2019hybrid}. 

\section{Datasets and evaluation}
\label{dataset}
\subsection{Datasets}
SemEval 2010 Task-8 Dataset \citep{hendrickx2009semeval} contains 2717 sentences, which do not overlap with the 8000 training instances from the version that was released on March 5, 2010 and the instances from SemEval 2007 Task-4. The dataset has 9 distinct relation type.

         NYT Dataset (NYT10) \cite{riedel2010modeling}, was created by aligning relations in Freebase with the sentences in the New York Times Annotated Corpus. Training and test set is generated by splitting the dataset by specific years. 
         Numerous previous work used NYT dataset for relation extraction tasks, however they leverage the dataset as their option. 
         
        FewRel \citep{gao2019fewrel} is a supervised dataset created to be used in relation classification methods which utilize few-shot approaches. Large set of sentences were first assigned to relations via distant supervision, next, they were annotated by human experts for denoising. The dataset contains 100 relations, each of which has 700 instances. 

         Wiki80 is created based on FewRel dataset for few-shot relation extraction tasks, however it is not recognized as a benchmark. It consists of 56,000 samples of 80 distinct relations. The samples are gathered from Wikidata and Wikipedia.

        TACRED is the crowd-annotated TAC Relation Extraction Dataset developed by The Stanford NLP Group \citep{zhang2017position}. 
        TACRED contains 106,264 samples and 41 relation types with ``no\_relation'' label to indicate that there is no relation between entities.

         ACE-2005 Multilingal Training Corpus is created for English, Chinese and Arabic languages \citep{walker2005ace} for the 2005 Automatic Content Extraction (ACE) technology evaluation. The datasets consist of various types of annotated data for entities, relations and events.

      WebNLG \citep{gardent-etal-2017-webnlg} is another dataset generated for NLP methods. \citet{zeng2018extracting} adapted this dataset for relation extraction tasks. The processed dataset contains 246 relation types, 5019 training, 703 test and 500 validation instances.

\subsection{Evaluation}

For supervised relation classification tasks the standard precision, recall and F-measure are used for evaluation. Authors usually provide the precision-recall curves for their classification results. For distantly supervised relation extraction models, held-out and/or manual evaluation is conducted. The labels of aligned text with a knowledge base are not gold. For this reason, only the relational facts coming from the knowledge base are considered to be true for the test set in held-out evaluation, newly predicted relations are treated as false. Since this assumption does not express the reality, some work (see Table \ref{table:3}) conduct manual evaluation which requires human effort. In few-shot learning, there are configurations in the form of $m$ way $n$ shot, $m$ representing the number of relations (classes), and $n$ representing labeled instance number per relation, in this case, sentences. The models are tested on different configuration of data and accuracy results of models on test set are stated.

\section{Discussion to solve common difficulties}
\label{ideas}
Neural relation extraction heavily makes use of the research on both deep learning and the semantic web. 
In this section, we discuss possible research directions regarding relation extraction.

\subsection{Question generation and question answering}
Neural question generation from text is an emerging research field \citep{zhou2017neural, yuan2017machine, kupiec1996method, du2017learning, wolfe1976automatic}. Question-answering on the knowledge graph is also a well-studied research topic \citep{rajpurkar2016squad}. 
Joint use of these studies on question generation from text and question answering on knowledge bases can help to discover missing relations between entities. Appropriate questions can be generated from each sentence using neural question generation methods. Each question is asked to the knowledge graph using question-answering methods that work on the knowledge base. If the system gets a response, then the response can be added to the natural text. Furthermore, a new triple can be generated based on the question and the response, which is appended to the existing triples in the knowledge base. As a result, the training data provides more insights in deep learning methods, as both the natural text and the knowledge graph is enhanced using question generation and question answering methods. 

\subsection{Possible solutions to improve results of attention mechanism}
In relation extraction, attention mechanism usually works best with sentences ranked according to how well they match a specific relation. 
A similarity metric is involved in matching the relation with the sentence. In this regard, various similarity methods can be explored. 
Based on the results of the relation extraction, importance weights can be refined, until the algorithm comes up with optimum results. 

Another possible approach for improvement is to run event detection algorithms on each of the sentences, paragraphs or documents, and make use of the events in sentence encoding and attention mechanism. 
Once the sentence's event is determined, event type-specific triples can be given a higher ranking to be aligned with the sentence. 
Event detection from a knowledge graph is also an emerging research area \citep{wang2019adversarial, wang2019hmeae}.

\subsection{Multilingual bi-text mining in machine translation}
One possible research field that can be integrated with distant supervision is machine translation using neural network models.
Machine translation models require a comprehensive training corpus which comprises aligned sentences in different languages. 
Consequently, sentence alignment is significant in machine translation. 

Google's Universal Sentence Encoder (USE) \citep{cer2018universal} embeds sentences into vectors by maintaining the context of the whole sentence, with a pre-trained model available for public use. 
An extension to USE which supports multilingual sentence encoding has also been published \cite{yang2019multilingual}. Facebook also published a similar multilingual sentence encoding study called Laser \cite{artetxe2019massively}. These studies make ``bi-text mining'' possible, which captures similarity scores of sentences even if they are in different languages and matches sentences having close similarity scores. 

Triples used in distant supervision can also be converted to sentences using natural language generation (NLG). Several studies exist for generating natural text from triples \cite{bouayad2014natural,duma2013generating,sun2006domain,cui-etal-2019-kb,triple_to_text}. Once triples are converted to natural text, the problem reduces to bi-text mining, where pre-trained models such as 
USE and Laser can be utilized. 
It is also possible to align knowledge bases and natural text in different languages using NLG and a multilingual sentence encoding tool. Since most of the general-purpose knowledge bases have more content in English, being able to align them with sentences in different languages can lead to tremendous improvement in distant supervision.

\subsection{Paraphrasing}
Another improvement in distant supervision can be achieved by expressing sentences in the natural text with different words.
Several paraphrased version of a sentence could be encoded into a vector space along with the original sentence. 
This can help to capture latent words and can result in a higher similarity score with related triples from knowledge bases.
One possible drawback of this approach could be that while producing abundant data, it simultaneously increases the number of false positives. 
The remedy could be extending the model with reinforcement learning or adversarial learning. 
As stated in section \ref{model-textension}, such methods give promising results in noise filtering. 

\subsection{Possible solutions for document-level relation extraction}
As stated in section \ref{document-re}, current methods on document-level relation extraction give poor results comparing to human performance. 
Enhancing the knowledge base and natural text can be relevant in this scenario, as it assists in finding hidden relation using neural relation prediction methods on the knowledge graph. Besides, external ontologies can be used to enhance the natural text, as ontologies include vocabularies and rule-sets. Locality-sensitive hashing (LSH) methods \citep{datar2004locality, aydar2019improved} could also be adopted to quickly determine which ontology aligns well with the input sentence, paragraph or document. 

\subsection{Integration with few-shot relation extraction}
Modern methods on neural relation extraction filter out instances that do not have a sufficient amount of training data. 
As stated in section \ref{fewshot-re}, few-shot relation extraction is suitable with small amount of training samples. In real scenarios, eliminating instances might not be desired. 
As a result, a joint method of neural relation extraction using distant supervision with a few-shot relation extraction algorithm might be more suitable for real-life scenarios.

\section{Conclusion}
\label{conclusion}
In this survey, we summarized neural relation extraction methods in terms of their approaches and data supervision and datasets for this task. In addition, we explained common challenges and discussed possible remedies to them.

To acquire abundant training instances, the latest studies make use of distant supervision. However, it brings noise to data which greatly affects the training of relation extraction models. In addition, there are no explicit negative samples, since the data itself have wrong annotations due to ill-alignment of the unstructured text and the knowledge graph. For that reason, instead of sentence-level approaches in supervised relation extraction, multi-instance approaches are developed for relation extraction with distant supervision. Also, few-shot learning for relation extraction is a research area that has still room for improvement. Supervised approaches are not to be abandoned. Indeed, incorporating pre-trained language models in supervised relation extraction make significant improvements in comparison to using conventional deep learning methods. Instead of treating entity recognition and relation extraction separately as in pipeline approaches, later studies adopt end-to-end approaches jointly extracting the entities and relations, which tend to better handle problems associated with overlapping triples and long-tail relations. 

\bibliography{references.bib}

\begin{thebibliography}{}

\bibitem[Artetxe and Schwenk, 2019]{artetxe2019massively}
Artetxe, M. and Schwenk, H. (2019).
\newblock Massively multilingual sentence embeddings for zero-shot
  cross-lingual transfer and beyond.
\newblock {\em Transactions of the Association for Computational Linguistics},
  7:597--610.

\bibitem[Auer et~al., 2007]{auer2007dbpedia}
Auer, S., Bizer, C., Kobilarov, G., Lehmann, J., Cyganiak, R., and Ives, Z.
  (2007).
\newblock Dbpedia: A nucleus for a web of open data.
\newblock In {\em The semantic web}, pages 722--735. Springer.

\bibitem[Aydar and Ayvaz, 2019]{aydar2019improved}
Aydar, M. and Ayvaz, S. (2019).
\newblock An improved method of locality-sensitive hashing for scalable
  instance matching.
\newblock {\em Knowledge and Information Systems}, 58(2):275--294.

\bibitem[Bollacker et~al., 2008]{bollacker2008freebase}
Bollacker, K., Evans, C., Paritosh, P., Sturge, T., and Taylor, J. (2008).
\newblock Freebase: a collaboratively created graph database for structuring
  human knowledge.
\newblock In {\em Proceedings of the 2008 ACM SIGMOD international conference
  on Management of data}, pages 1247--1250. AcM.

\bibitem[Bouayad-Agha et~al., 2014]{bouayad2014natural}
Bouayad-Agha, N., Casamayor, G., and Wanner, L. (2014).
\newblock Natural language generation in the context of the semantic web.
\newblock {\em Semantic Web}, 5(6):493--513.

\bibitem[Cai et~al., 2016]{cai2016bidirectional}
Cai, R., Zhang, X., and Wang, H. (2016).
\newblock Bidirectional recurrent convolutional neural network for relation
  classification.
\newblock In {\em Proceedings of the 54th Annual Meeting of the Association for
  Computational Linguistics (Volume 1: Long Papers)}, pages 756--765.

\bibitem[Cer et~al., 2018]{cer2018universal}
Cer, D., Yang, Y., Kong, S.-y., Hua, N., Limtiaco, N., John, R.~S., Constant,
  N., Guajardo-Cespedes, M., Yuan, S., Tar, C., et~al. (2018).
\newblock Universal sentence encoder.
\newblock {\em arXiv preprint arXiv:1803.11175}.

\bibitem[Cui et~al., 2019]{cui-etal-2019-kb}
Cui, W., Zhou, M., Zhao, R., and Norouzi, N. (2019).
\newblock {KB}-{NLG}: From knowledge base to natural language generation.
\newblock In {\em Proceedings of the 2019 Workshop on Widening NLP}, pages
  80--82, Florence, Italy. Association for Computational Linguistics.

\bibitem[Dai et~al., 2019]{dai2019transformer}
Dai, Z., Yang, Z., Yang, Y., Carbonell, J.~G., Le, Q.~V., and Salakhutdinov, R.
  (2019).
\newblock Transformer-xl: Attentive language models beyond a fixed-length
  context.
\newblock {\em CoRR}, abs/1901.02860.

\bibitem[Datar et~al., 2004]{datar2004locality}
Datar, M., Immorlica, N., Indyk, P., and Mirrokni, V.~S. (2004).
\newblock Locality-sensitive hashing scheme based on p-stable distributions.
\newblock In {\em Proceedings of the twentieth annual symposium on
  Computational geometry}, pages 253--262. ACM.

\bibitem[Devlin et~al., 2018]{devlin2018bert}
Devlin, J., Chang, M.-W., Lee, K., and Toutanova, K. (2018).
\newblock Bert: Pre-training of deep bidirectional transformers for language
  understanding.
\newblock {\em arXiv preprint arXiv:1810.04805}.

\bibitem[Dietterich et~al., 1997]{dietterich1997solving}
Dietterich, T.~G., Lathrop, R.~H., and Lozano-P{\'e}rez, T. (1997).
\newblock Solving the multiple instance problem with axis-parallel rectangles.
\newblock {\em Artificial intelligence}, 89(1-2):31--71.

\bibitem[Du et~al., 2017]{du2017learning}
Du, X., Shao, J., and Cardie, C. (2017).
\newblock Learning to ask: Neural question generation for reading
  comprehension.
\newblock {\em arXiv preprint arXiv:1705.00106}.

\bibitem[Duma and Klein, 2013]{duma2013generating}
Duma, D. and Klein, E. (2013).
\newblock Generating natural language from linked data: Unsupervised template
  extraction.
\newblock In {\em Proceedings of the 10th International Conference on
  Computational Semantics (IWCS 2013)--Long Papers}, pages 83--94.

\bibitem[Feng et~al., 2018]{feng2018reinforcement}
Feng, J., Huang, M., Zhao, L., Yang, Y., and Zhu, X. (2018).
\newblock Reinforcement learning for relation classification from noisy data.
\newblock In {\em Thirty-Second AAAI Conference on Artificial Intelligence}.

\bibitem[Fu et~al., 2019]{fu2019graphrel}
Fu, T.-J., Li, P.-H., and Ma, W.-Y. (2019).
\newblock Graphrel: Modeling text as relational graphs for joint entity and
  relation extraction.
\newblock In {\em Proceedings of the 57th Annual Meeting of the Association for
  Computational Linguistics}, pages 1409--1418.

\bibitem[Gao et~al., 2019a]{gao2019hybrid}
Gao, T., Han, X., Liu, Z., and Sun, M. (2019a).
\newblock Hybrid attention-based prototypical networks for noisy few-shot
  relation classification.

\bibitem[Gao et~al., 2019b]{gao2019fewrel}
Gao, T., Han, X., Zhu, H., Liu, Z., Li, P., Sun, M., and Zhou, J. (2019b).
\newblock Fewrel 2.0: Towards more challenging few-shot relation
  classification.
\newblock {\em arXiv preprint arXiv:1910.07124}.

\bibitem[Gardent et~al., 2017]{gardent-etal-2017-webnlg}
Gardent, C., Shimorina, A., Narayan, S., and Perez-Beltrachini, L. (2017).
\newblock The {W}eb{NLG} challenge: Generating text from {RDF} data.
\newblock In {\em Proceedings of the 10th International Conference on Natural
  Language Generation}, pages 124--133, Santiago de Compostela, Spain.
  Association for Computational Linguistics.

\bibitem[Han et~al., 2019]{han2019opennre}
Han, X., Gao, T., Yao, Y., Ye, D., Liu, Z., and Sun, M. (2019).
\newblock Opennre: An open and extensible toolkit for neural relation
  extraction.
\newblock {\em arXiv preprint arXiv:1909.13078}.

\bibitem[Han et~al., 2018a]{han2018neural}
Han, X., Liu, Z., and Sun, M. (2018a).
\newblock Neural knowledge acquisition via mutual attention between knowledge
  graph and text.
\newblock In {\em Thirty-Second AAAI Conference on Artificial Intelligence}.

\bibitem[Han et~al., 2018b]{han2018hierarchical}
Han, X., Yu, P., Liu, Z., Sun, M., and Li, P. (2018b).
\newblock Hierarchical relation extraction with coarse-to-fine grained
  attention.
\newblock In {\em Proceedings of the 2018 Conference on Empirical Methods in
  Natural Language Processing}, pages 2236--2245.

\bibitem[Han et~al., 2018c]{han2018fewrel}
Han, X., Zhu, H., Yu, P., Wang, Z., Yao, Y., Liu, Z., and Sun, M. (2018c).
\newblock Fewrel: A large-scale supervised few-shot relation classification
  dataset with state-of-the-art evaluation.
\newblock In {\em Proceedings of the 2018 Conference on Empirical Methods in
  Natural Language Processing}, pages 4803--4809.

\bibitem[He et~al., 2016]{he2016deep}
He, K., Zhang, X., Ren, S., and Sun, J. (2016).
\newblock Deep residual learning for image recognition.
\newblock In {\em Proceedings of the IEEE conference on computer vision and
  pattern recognition}, pages 770--778.

\bibitem[Hendrickx et~al., 2009]{hendrickx2009semeval}
Hendrickx, I., Kim, S.~N., Kozareva, Z., Nakov, P., {\'O}~S{\'e}aghdha, D.,
  Pad{\'o}, S., Pennacchiotti, M., Romano, L., and Szpakowicz, S. (2009).
\newblock Semeval-2010 task 8: Multi-way classification of semantic relations
  between pairs of nominals.
\newblock In {\em Proceedings of the Workshop on Semantic Evaluations: Recent
  Achievements and Future Directions}, pages 94--99. Association for
  Computational Linguistics.

\bibitem[Hoffmann et~al., 2011]{hoffmann2011knowledge}
Hoffmann, R., Zhang, C., Ling, X., Zettlemoyer, L., and Weld, D.~S. (2011).
\newblock Knowledge-based weak supervision for information extraction of
  overlapping relations.
\newblock In {\em Proceedings of the 49th Annual Meeting of the Association for
  Computational Linguistics: Human Language Technologies}, pages 541--550,
  Portland, Oregon, USA. Association for Computational Linguistics.

\bibitem[Huang and Wang, 2017]{huang2017deep}
Huang, Y. and Wang, W.~Y. (2017).
\newblock Deep residual learning for weakly-supervised relation extraction.
\newblock In {\em Proceedings of the 2017 Conference on Empirical Methods in
  Natural Language Processing}, pages 1803--1807.

\bibitem[Ji et~al., 2017]{ji2017distant}
Ji, G., Liu, K., He, S., and Zhao, J. (2017).
\newblock Distant supervision for relation extraction with sentence-level
  attention and entity descriptions.
\newblock In {\em Thirty-First AAAI Conference on Artificial Intelligence}.

\bibitem[Jiang et~al., 2016]{jiang2016relation}
Jiang, X., Wang, Q., Li, P., and Wang, B. (2016).
\newblock Relation extraction with multi-instance multi-label convolutional
  neural networks.
\newblock In {\em Proceedings of COLING 2016, the 26th International Conference
  on Computational Linguistics: Technical Papers}, pages 1471--1480.

\bibitem[Kumar, 2017]{kumar2017survey}
Kumar, S. (2017).
\newblock A survey of deep learning methods for relation extraction.
\newblock {\em arXiv preprint arXiv:1705.03645}.

\bibitem[Kupiec, 1996]{kupiec1996method}
Kupiec, J.~M. (1996).
\newblock Method for extracting from a text corpus answers to questions stated
  in natural language by using linguistic analysis and hypothesis generation.
\newblock US Patent 5,519,608.

\bibitem[Lin et~al., 2017]{lin2017neural}
Lin, Y., Liu, Z., and Sun, M. (2017).
\newblock Neural relation extraction with multi-lingual attention.
\newblock In {\em Proceedings of the 55th Annual Meeting of the Association for
  Computational Linguistics (Volume 1: Long Papers)}, pages 34--43.

\bibitem[Lin et~al., 2016]{lin2016neural}
Lin, Y., Shen, S., Liu, Z., Luan, H., and Sun, M. (2016).
\newblock Neural relation extraction with selective attention over instances.
\newblock In {\em Proceedings of the 54th Annual Meeting of the Association for
  Computational Linguistics (Volume 1: Long Papers)}, pages 2124--2133.

\bibitem[Liu et~al., 2016]{liu2016effective}
Liu, A., Soderland, S., Bragg, J., Lin, C.~H., Ling, X., and Weld, D.~S.
  (2016).
\newblock Effective crowd annotation for relation extraction.
\newblock In {\em Proceedings of the 2016 Conference of the North {A}merican
  Chapter of the Association for Computational Linguistics: Human Language
  Technologies}, pages 897--906.

\bibitem[Liu et~al., 2017]{liu2017soft}
Liu, T., Wang, K., Chang, B., and Sui, Z. (2017).
\newblock A soft-label method for noise-tolerant distantly supervised relation
  extraction.
\newblock In {\em Proceedings of the 2017 Conference on Empirical Methods in
  Natural Language Processing}, pages 1790--1795.

\bibitem[Mintz et~al., 2009]{mintz2009distant}
Mintz, M., Bills, S., Snow, R., and Jurafsky, D. (2009).
\newblock Distant supervision for relation extraction without labeled data.
\newblock In {\em Proceedings of the Joint Conference of the 47th Annual
  Meeting of the ACL and the 4th International Joint Conference on Natural
  Language Processing of the AFNLP: Volume 2-Volume 2}, pages 1003--1011.
  Association for Computational Linguistics.

\bibitem[Nguyen and Grishman, 2015]{nguyen2015relation}
Nguyen, T.~H. and Grishman, R. (2015).
\newblock Relation extraction: Perspective from convolutional neural networks.
\newblock In {\em Proceedings of the 1st Workshop on Vector Space Modeling for
  Natural Language Processing}, pages 39--48, Denver, Colorado. Association for
  Computational Linguistics.

\bibitem[Pawar et~al., 2017]{pawar2017relation}
Pawar, S., Palshikar, G.~K., and Bhattacharyya, P. (2017).
\newblock Relation extraction: A survey.
\newblock {\em arXiv preprint arXiv:1712.05191}.

\bibitem[Qin et~al., 2018a]{qin2018dsgan}
Qin, P., Xu, W., and Wang, W.~Y. (2018a).
\newblock Dsgan: generative adversarial training for distant supervision
  relation extraction.
\newblock {\em arXiv preprint arXiv:1805.09929}.

\bibitem[Qin et~al., 2018b]{qin2018robust}
Qin, P., Xu, W., and Wang, W.~Y. (2018b).
\newblock Robust distant supervision relation extraction via deep reinforcement
  learning.
\newblock {\em arXiv preprint arXiv:1805.09927}.

\bibitem[Quirk and Poon, 2017]{quirk2017distant}
Quirk, C. and Poon, H. (2017).
\newblock Distant supervision for relation extraction beyond the sentence
  boundary.
\newblock In {\em Proceedings of the 15th Conference of the European Chapter of
  the Association for Computational Linguistics: Volume 1, Long Papers}, pages
  1171--1182.

\bibitem[Radford et~al., 2019]{radford2019language}
Radford, A., Wu, J., Child, R., Luan, D., Amodei, D., and Sutskever, I. (2019).
\newblock Language models are unsupervised multitask learners.
\newblock {\em OpenAI Blog}, 1(8).

\bibitem[Rajpurkar et~al., 2016]{rajpurkar2016squad}
Rajpurkar, P., Zhang, J., Lopyrev, K., and Liang, P. (2016).
\newblock Squad: 100,000+ questions for machine comprehension of text.
\newblock {\em arXiv preprint arXiv:1606.05250}.

\bibitem[Ratner et~al., 2016]{ratner2016data}
Ratner, A.~J., De~Sa, C.~M., Wu, S., Selsam, D., and R{\'e}, C. (2016).
\newblock Data programming: Creating large training sets, quickly.
\newblock In {\em Advances in neural information processing systems}, pages
  3567--3575.

\bibitem[Riedel et~al., 2010]{riedel2010modeling}
Riedel, S., Yao, L., and McCallum, A. (2010).
\newblock Modeling relations and their mentions without labeled text.
\newblock In {\em Joint European Conference on Machine Learning and Knowledge
  Discovery in Databases}, pages 148--163. Springer.

\bibitem[Santos et~al., 2015]{santos2015classifying}
Santos, C. N.~d., Xiang, B., and Zhou, B. (2015).
\newblock Classifying relations by ranking with convolutional neural networks.
\newblock {\em arXiv preprint arXiv:1504.06580}.

\bibitem[Smirnova and Cudr{\'e}-Mauroux, 2018]{smirnova2018relation}
Smirnova, A. and Cudr{\'e}-Mauroux, P. (2018).
\newblock Relation extraction using distant supervision: A survey.
\newblock {\em ACM Computing Surveys (CSUR)}, 51(5):106.

\bibitem[Snell et~al., 2017]{snell2017prototypical}
Snell, J., Swersky, K., and Zemel, R. (2017).
\newblock Prototypical networks for few-shot learning.
\newblock In {\em Advances in Neural Information Processing Systems}, pages
  4077--4087.

\bibitem[Soares et~al., 2019]{soares2019matching}
Soares, L.~B., FitzGerald, N., Ling, J., and Kwiatkowski, T. (2019).
\newblock Matching the blanks: Distributional similarity for relation learning.
\newblock {\em arXiv preprint arXiv:1906.03158}.

\bibitem[Socher et~al., 2012]{socher2012semantic}
Socher, R., Huval, B., Manning, C.~D., and Ng, A.~Y. (2012).
\newblock Semantic compositionality through recursive matrix-vector spaces.
\newblock In {\em Proceedings of the 2012 joint conference on empirical methods
  in natural language processing and computational natural language learning},
  pages 1201--1211. Association for Computational Linguistics.

\bibitem[Sun and Mellish, 2006]{sun2006domain}
Sun, X. and Mellish, C. (2006).
\newblock Domain independent sentence generation from rdf representations for
  the semantic web.
\newblock In {\em Combined Workshop on Language-Enabled Educational Technology
  and Development and Evaluation of Robust Spoken Dialogue Systems, European
  Conference on AI, Riva del Garda, Italy}.

\bibitem[Surdeanu et~al., 2012]{surdeanu2012multi}
Surdeanu, M., Tibshirani, J., Nallapati, R., and Manning, C.~D. (2012).
\newblock Multi-instance multi-label learning for relation extraction.
\newblock In {\em Proceedings of the 2012 Joint Conference on Empirical Methods
  in Natural Language Processing and Computational Natural Language Learning},
  pages 455--465, Jeju Island, Korea. Association for Computational
  Linguistics.

\bibitem[Takamatsu et~al., 2012]{takamatsu2012reducing}
Takamatsu, S., Sato, I., and Nakagawa, H. (2012).
\newblock Reducing wrong labels in distant supervision for relation extraction.
\newblock In {\em Proceedings of the 50th Annual Meeting of the Association for
  Computational Linguistics: Long Papers-Volume 1}, pages 721--729. Association
  for Computational Linguistics.

\bibitem[Takanobu et~al., 2019]{takanobu2019hierarchical}
Takanobu, R., Zhang, T., Liu, J., and Huang, M. (2019).
\newblock A hierarchical framework for relation extraction with reinforcement
  learning.
\newblock In {\em Proceedings of the AAAI Conference on Artificial
  Intelligence}, volume~33, pages 7072--7079.

\bibitem[Verga et~al., 2018]{verga2018simultaneously}
Verga, P., Strubell, E., and McCallum, A. (2018).
\newblock Simultaneously self-attending to all mentions for full-abstract
  biological relation extraction.
\newblock In {\em Proceedings of the 2018 Conference of the North {A}merican
  Chapter of the Association for Computational Linguistics: Human Language
  Technologies, Volume 1 (Long Papers)}, pages 872--884, New Orleans,
  Louisiana. Association for Computational Linguistics.

\bibitem[Walker et~al., 2005]{walker2005ace}
Walker, C., Strassel, S., Medero, J., and Maeda, K. (2005).
\newblock Ace 2005 multilingual training corpus-linguistic data consortium.
\newblock {\em URL: https://catalog. ldc. upenn. edu/LDC2006T06}.

\bibitem[Wang et~al., 2018a]{wang2018label}
Wang, G., Zhang, W., Wang, R., Zhou, Y., Chen, X., Zhang, W., Zhu, H., and
  Chen, H. (2018a).
\newblock Label-free distant supervision for relation extraction via knowledge
  graph embedding.
\newblock In {\em Proceedings of the 2018 Conference on Empirical Methods in
  Natural Language Processing}, pages 2246--2255.

\bibitem[Wang et~al., 2016]{wang2016relation}
Wang, L., Cao, Z., De~Melo, G., and Liu, Z. (2016).
\newblock Relation classification via multi-level attention cnns.
\newblock In {\em Proceedings of the 54th annual meeting of the Association for
  Computational Linguistics (volume 1: long papers)}, pages 1298--1307.

\bibitem[Wang et~al., 2018b]{wang2018adversarial}
Wang, X., Han, X., Lin, Y., Liu, Z., and Sun, M. (2018b).
\newblock Adversarial multi-lingual neural relation extraction.
\newblock In {\em Proceedings of the 27th International Conference on
  Computational Linguistics}, pages 1156--1166.

\bibitem[Wang et~al., 2019a]{wang2019adversarial}
Wang, X., Han, X., Liu, Z., Sun, M., and Li, P. (2019a).
\newblock Adversarial training for weakly supervised event detection.
\newblock In {\em Proceedings of the 2019 Conference of the North American
  Chapter of the Association for Computational Linguistics: Human Language
  Technologies, Volume 1 (Long and Short Papers)}, pages 998--1008.

\bibitem[Wang et~al., 2019b]{wang2019hmeae}
Wang, X., Wang, Z., Han, X., Liu, Z., Li, J., Li, P., Sun, M., Zhou, J., and
  Ren, X. (2019b).
\newblock {HMEAE}: Hierarchical modular event argument extraction.
\newblock In {\em Proceedings of the 2019 Conference on Empirical Methods in
  Natural Language Processing and the 9th International Joint Conference on
  Natural Language Processing (EMNLP-IJCNLP)}, pages 5781--5787, Hong Kong,
  China. Association for Computational Linguistics.

\bibitem[Wei et~al., 2019]{wei2019novel}
Wei, Z., Su, J., Wang, Y., Tian, Y., and Chang, Y. (2019).
\newblock A novel hierarchical binary tagging framework for joint extraction of
  entities and relations.
\newblock {\em arXiv preprint arXiv:1909.03227}.

\bibitem[Wolfe, 1976]{wolfe1976automatic}
Wolfe, J.~H. (1976).
\newblock Automatic question generation from text-an aid to independent study.
\newblock In {\em ACM SIGCUE Outlook}, volume~10, pages 104--112. ACM.

\bibitem[Wu and He, 2019]{wu2019enriching}
Wu, S. and He, Y. (2019).
\newblock Enriching pre-trained language model with entity information for
  relation classification.
\newblock {\em arXiv preprint arXiv:1905.08284}.

\bibitem[Wu et~al., 2017]{wu2017adversarial}
Wu, Y., Bamman, D., and Russell, S. (2017).
\newblock Adversarial training for relation extraction.
\newblock In {\em Proceedings of the 2017 Conference on Empirical Methods in
  Natural Language Processing}, pages 1778--1783.

\bibitem[Xu et~al., 2015]{xu2015classifying}
Xu, Y., Mou, L., Li, G., Chen, Y., Peng, H., and Jin, Z. (2015).
\newblock Classifying relations via long short term memory networks along
  shortest dependency paths.
\newblock In {\em proceedings of the 2015 conference on empirical methods in
  natural language processing}, pages 1785--1794.

\bibitem[Yang et~al., 2019]{yang2019multilingual}
Yang, Y., Cer, D., Ahmad, A., Guo, M., Law, J., Constant, N., Abrego, G.~H.,
  Yuan, S., Tar, C., Sung, Y.-H., et~al. (2019).
\newblock Multilingual universal sentence encoder for semantic retrieval.
\newblock {\em arXiv preprint arXiv:1907.04307}.

\bibitem[Yao et~al., 2019]{yao2019docred}
Yao, Y., Ye, D., Li, P., Han, X., Lin, Y., Liu, Z., Liu, Z., Huang, L., Zhou,
  J., and Sun, M. (2019).
\newblock Docred: A large-scale document-level relation extraction dataset.
\newblock {\em arXiv preprint arXiv:1906.06127}.

\bibitem[Ye and Ling, 2019]{ye2019multi}
Ye, Z.-X. and Ling, Z.-H. (2019).
\newblock Multi-level matching and aggregation network for few-shot relation
  classification.
\newblock {\em arXiv preprint arXiv:1906.06678}.

\bibitem[Yuan et~al., 2017]{yuan2017machine}
Yuan, X., Wang, T., Gulcehre, C., Sordoni, A., Bachman, P., Subramanian, S.,
  Zhang, S., and Trischler, A. (2017).
\newblock Machine comprehension by text-to-text neural question generation.
\newblock {\em arXiv preprint arXiv:1705.02012}.

\bibitem[Zeng et~al., 2015]{zeng2015distant}
Zeng, D., Liu, K., Chen, Y., and Zhao, J. (2015).
\newblock Distant supervision for relation extraction via piecewise
  convolutional neural networks.
\newblock In {\em Proceedings of the 2015 Conference on Empirical Methods in
  Natural Language Processing}, pages 1753--1762.

\bibitem[Zeng et~al., 2014]{zeng2014relation}
Zeng, D., Liu, K., Lai, S., Zhou, G., Zhao, J., et~al. (2014).
\newblock Relation classification via convolutional deep neural network.

\bibitem[Zeng et~al., 2017]{zeng2017incorporating}
Zeng, W., Lin, Y., Liu, Z., and Sun, M. (2017).
\newblock Incorporating relation paths in neural relation extraction.
\newblock In {\em Proceedings of the 2017 Conference on Empirical Methods in
  Natural Language Processing}, pages 1768--1777.

\bibitem[Zeng et~al., 2018]{zeng2018extracting}
Zeng, X., Zeng, D., He, S., Liu, K., Zhao, J., et~al. (2018).
\newblock Extracting relational facts by an end-to-end neural model with copy
  mechanism.

\bibitem[Zhang and Wang, 2015]{zhang2015relation}
Zhang, D. and Wang, D. (2015).
\newblock Relation classification via recurrent neural network.
\newblock {\em arXiv preprint arXiv:1508.01006}.

\bibitem[Zhang et~al., 2015]{zhang2015bidirectional}
Zhang, S., Zheng, D., Hu, X., and Yang, M. (2015).
\newblock Bidirectional long short-term memory networks for relation
  classification.
\newblock In {\em Proceedings of the 29th Pacific Asia Conference on Language,
  Information and Computation}, pages 73--78, Shanghai, China.

\bibitem[Zhang et~al., 2017]{zhang2017position}
Zhang, Y., Zhong, V., Chen, D., Angeli, G., and Manning, C.~D. (2017).
\newblock Position-aware attention and supervised data improve slot filling.
\newblock In {\em Proceedings of the 2017 Conference on Empirical Methods in
  Natural Language Processing}, pages 35--45.

\bibitem[Zhao et~al., 2019]{zhao2019improving}
Zhao, Y., Wan, H., Gao, J., and Lin, Y. (2019).
\newblock Improving relation classification by entity pair graph.
\newblock In {\em Asian Conference on Machine Learning}, pages 1156--1171.

\bibitem[Zheng et~al., 2019]{zheng2019diag}
Zheng, S., Han, X., Lin, Y., Yu, P., Chen, L., Huang, L., Liu, Z., and Xu, W.
  (2019).
\newblock Diag-nre: A neural pattern diagnosis framework for distantly
  supervised neural relation extraction.
\newblock In {\em Proceedings of the 57th Annual Meeting of the Association for
  Computational Linguistics}, pages 1419--1429.

\bibitem[Zhou et~al., 2016]{zhou2016attention}
Zhou, P., Shi, W., Tian, J., Qi, Z., Li, B., Hao, H., and Xu, B. (2016).
\newblock Attention-based bidirectional long short-term memory networks for
  relation classification.
\newblock In {\em Proceedings of the 54th Annual Meeting of the Association for
  Computational Linguistics (Volume 2: Short Papers)}, pages 207--212.

\bibitem[Zhou et~al., 2017]{zhou2017neural}
Zhou, Q., Yang, N., Wei, F., Tan, C., Bao, H., and Zhou, M. (2017).
\newblock Neural question generation from text: A preliminary study.
\newblock In {\em National CCF Conference on Natural Language Processing and
  Chinese Computing}, pages 662--671. Springer.

\bibitem[Zhu et~al., 2019]{triple_to_text}
Zhu, Y., Wan, J., Zhou, Z., Chen, L., Qiu, L., Zhang, W., Jiang, X., and Yu, Y.
  (2019).
\newblock Triple-to-text: Converting {RDF} triples into high-quality natural
  languages via optimizing an inverse {KL} divergence.
\newblock {\em CoRR}, abs/1906.01965.

\end{thebibliography}
\bibliographystyle{apalike}

% \newpage
% \clearpage

% \appendix
% \renewcommand{\thesection}{\Alph{section}}
% \input{supplement}

\end{document}